\documentclass[preprint]{elsarticle}
\usepackage{hyperref}
\usepackage{amsmath,amssymb,amsfonts}
\usepackage{algorithmic}
\usepackage{graphicx}
\usepackage{textcomp}
\usepackage{subcaption} 
\usepackage{caption,setspace}
\usepackage{tabto}
\usepackage{multirow}
\usepackage{mathtools}
\DeclarePairedDelimiter\floor{\lfloor}{\rfloor}
\journal{JKSUCI, Elsevier (Accepted Manuscript).}

\begin{document}

\begin{frontmatter}

\title{BDNet: Bengali Handwritten Numeral Digit Recognition based on Densely connected Convolutional Neural Networks}


\author{Abu Sufian\fnref{myfootnote} \corref{mycorrespondingauthor}}
\cortext[mycorrespondingauthor]{Corresponding author}
\ead{sufian.csa@gmail.com}
\author{Anirudha Ghosh\fnref{myfootnote}}
\author{Avijit Naskar\fnref{myfootnote}}
\author{Farhana Sultana\fnref{myfootnote}}
\author{Jaya Sil\fnref{myfootnote1}}
\author{M M Hafizur Rahman\fnref{myfootnote2}}
\fntext[myfootnote]{Dept. of Computer Science, University of Gour Banga, West Bengal, India.}
\fntext[myfootnote1]{Dept. of Computer Science \& Technology, IIEST Shibpur, West Bengal, India.}
\fntext[myfootnote2]{Dept. of Computer Networks and Communications, CCSIT, King Faisal University, Al Ahsa 31982, Saudi Arabia.}




\begin{abstract}
Images of handwritten digits are different from natural images as the orientation of a digit, as well as similarity of features of different digits, makes confusion. On the other hand, deep convolutional neural networks are achieving huge success in computer vision problems, especially in image classification. Here, we propose a task-oriented model called Bengali handwritten numeral digit recognition based on densely connected convolutional neural networks(BDNet). BDNet is used to classify (recognize) Bengali handwritten numeral digits. It is end-to-end trained using ISI Bengali handwritten numeral dataset. During training, untraditional data preprocessing and augmentation techniques are used so that the trained model works on a different dataset. The model has achieved the test accuracy of \textbf{99.78\%}(baseline was 99.58\%) on the test dataset of ISI Bengali handwritten numerals. So, the BDNet model gives 47.62\% error reduction compared to previous state-of-the-art models.  Here we have also created a dataset of 1000 images of Bengali handwritten numerals to test the trained model, and it giving promising results. Codes, trained model and our own dataset are available at:\href{https://github.com/Sufianlab/BDNet}C.   
\end{abstract}

\begin{keyword}
Bengali Digit Recognition\sep CNN\sep Dataset \sep Deep Learning\sep Handwritten Numerals \sep Image Classification.
\end{keyword}

\end{frontmatter}


\section{Introduction}
Bangla (Bengali) is the second most spoken language in India. It ranks fifth in Asia and it is also in the top ten spoken languages in the world \cite{prasenjit07}. So, a huge number of people depend on this language for their day to day communication. Therefore, automatic recognition of Bengali handwritten characters and numeral digits are needed to be digitized for making the communication smoother. Many research works and models have been proposed to recognize Bengali handwritten characters and numeral digits so far, but still, a huge scope is there to improve this task in terms of accuracy and applicability. Most of the previously proposed models are based on traditional pattern recognition and machine learning techniques where human expertise is required for feature engineering \cite{BAG2013}, \cite{umapada12}. \par   
\begin{figure}[htb]
	\centering
	\includegraphics[width=.65\linewidth]{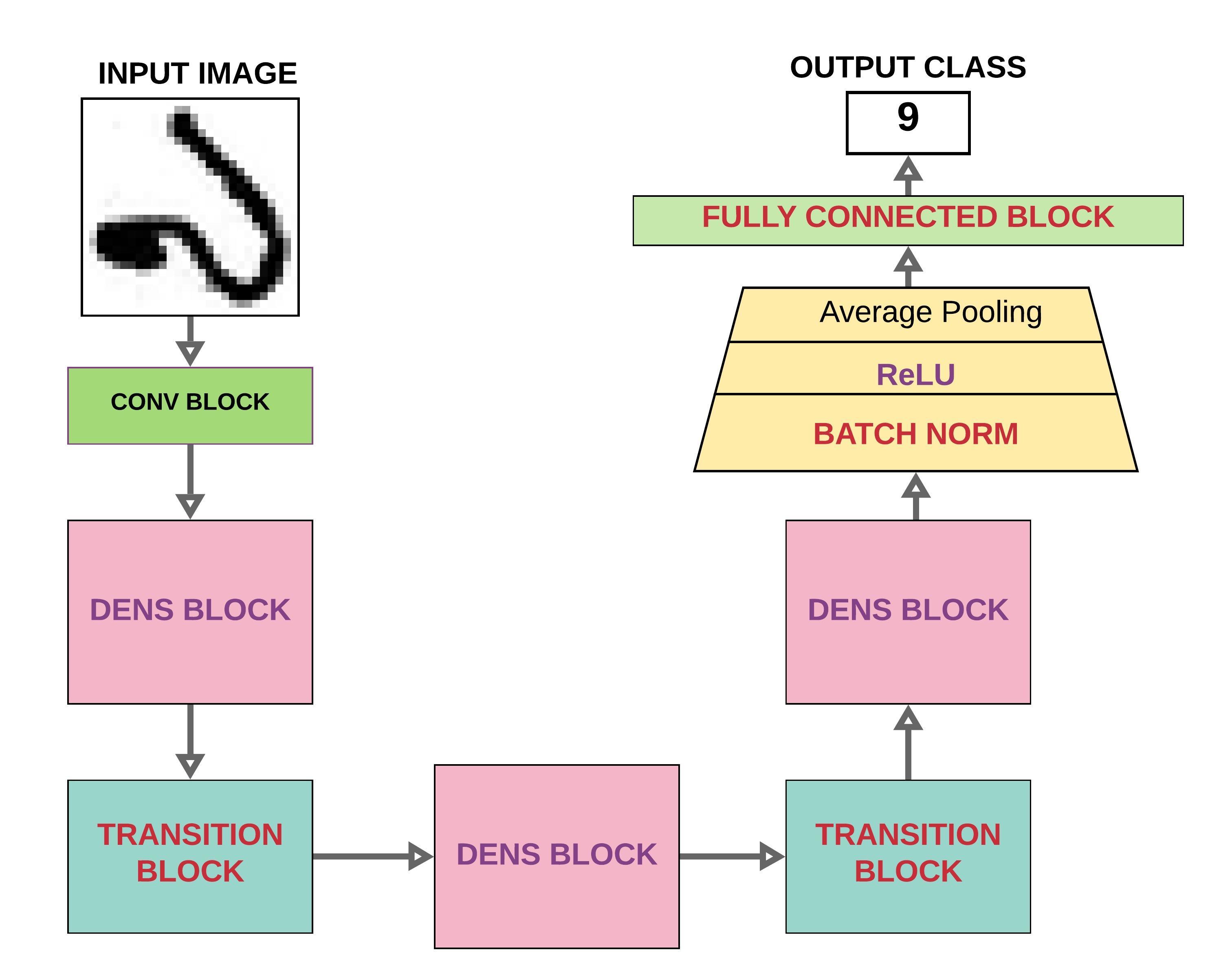}
	\caption{Overview of the BDNet model.}
	\label{pipeline}
\end{figure}
The recent success of deep learning, especially Convolutional Neural Network (CNN) for computer vision \cite{CNN19}, \cite{alexnet12}, \cite{yann15}, \cite{wang18} \cite{farhana18}  has inspired many researchers to use the CNN to recognize handwritten characters and digits as a computer vision task. The BDNet, we proposing through this paper, is a task-oriented deep CNN based model to classify Bengali numeral digits. This model trained using untraditional preprocessing and data augmentation techniques for trained with generalization.  It has achieved new baseline accuracy on benchmark datasets \cite{ujjwal09}. Here we have also proposed a new dataset of 1000 images of Bengali handwritten numerals. This own dataset used to test the trained model, and it giving promising results but this new dataset not used during training.  The working pipeline of the BDNet is designed by an inspiration of DenseNet \cite{densenet2017} which is one of the state-of-the-art deep CNN algorithm for image classification. The  conceptual view of the BDNet has shown in figure \ref{pipeline} and details of the model have explained in section \ref{Model}.

\subsection{Contributions of this paper}
\begin{itemize}
	\item A deep CNN working model, called BDNet, is designed to recognize Bengali handwritten numerals.
	\item  The BDNet is trained end-to-end with unorthodox data preprocessing and augmentation.  
	\item The proposed model has achieved the highest test accuracy with 47.62\% error reduction on baseline result(\textbf{99.78\%} whereas the previous best was 99.58\%) on the test data of ISI Bengali handwritten numeral dataset.
	\item A new dataset is created with 1000 samples of handwritten Bengali numerals for testing performance of the trained BDNet where the model gives promising results. 
\end{itemize}
\subsection{Organization of the paper}
The rest of the paper is organized as follows: In section \ref{review}, a literature review has done. Model details of BDNet are explained in section \ref{Model}. In section \ref{dataset}, the dataset and preprocessing of data are explained. Training details are explained in section \ref{traning} and result analysis is in section \ref{result}. Finally, the paper is concluded in section \ref{conclusion}.      
\section{Literature Review}
\label{review}
For background analysis, we have reviewed two relevant things: one is existing research works on Bengali handwritten numeral recognition and another is the advancements of deep learning models for image classification. The first part of this review is for baseline results and domain knowledge of target application, whereas later part is for the idea about the latest trends of deep learning-based state-of-the-art algorithms. In this section, we have reviewed these two things in the following two subsections:
\subsection{Existing works on Bengali handwritten numeral recognition}
\label{exixting works}
Bengali handwritten numeral recognition is one of the oldest pattern recognition problems. Many researchers have been working in this field since the 90s of the last century \cite{abhijit93}, \cite{umapada00}. Through this subsection, we have reviewed most notable works on this Bengali handwritten numeral recognition.\par 
Subhadip Basu et al. proposed Handwritten Bangla Digit Recognition using Classifier Combination Through Dempster-Shafer (DS) Technique \cite{basu05}. They have used the DS technique and MLP classifier for classification and also used 3-fold cross-validation on the training dataset of 6000 handwritten samples. Their scheme achieved 95.1\% test accuracy. In \cite{umapada06} U. Pal et al. proposed a scheme where unconstrained off-line Bengali handwritten numerals were recognized. This scheme has recognized different handwritten styles. The scheme selects the required features using the concept of water overflow from the reservoir, and also collect topological and structural features of the numerals. They applied this scheme on their own collected dataset of size 12000 and obtained recognition accuracy of around 92.8\%. \par 
U. Bhattacharya and B. B. Choudhury presented a handwritten numeral database along with the Devanagari database and proposed a classifier model \cite{ujjwal09}. Their database contains 23392 handwritten Bengali numeral images. Their classifier model is a multi-stage cascaded recognition scheme where they used wavelet-based multi-resolution representations and multilayer perception as classifiers. They have mentioned 99.14\% training and 98.20\% testing accuracy on this dataset. Cheng-Lin Liu and Ching Y. Suen proposed a benchmark model \cite{liu09} on the ISI numeral dataset \cite{ujjwal09} along with a Farsi numeral database. They preprocessed the dataset into gray-scale images and applied many traditional feature extraction models. This benchmark model achieved the highest test accuracy of 99.40\%. Ying Wen and Lianghua He proposed a classifier model \cite{wen12} for Bengali handwritten numeral recognition. This model tried to solve large dataset high dimensionality problem. They combined Bayesian discriminant with kernel approach with UCI dataset and another dataset such as MNIST \cite{yann_mnist}. The rate of error is 1.8\%, the recognition rate is 99.08\% and recognition time is 7.46 milliseconds. \par
Local region identification, where optimal unambiguous features are extracted, is one of the crucial tasks in the field of character recognition. This idea is adopted by N. Das et al. in their handwritten digit recognition technique \cite{nibaran12} based on a genetic algorithm(GA). GA is applied to seven sets of local regions. For each set, GA selects a minimal local region group with a Support Vector Machine (SVM) based classifier. The whole digit images are used for global feature extraction whereas local features are extracted for shape information. The number of global features is constant whereas the number of local features depends on the number of the local regions. The test accuracy rate was 95.50\% for this model.
M. K. Nasir and M. S. Uddin proposed a scheme \cite{Nasir13} where they used K-Means clustering, Bayes' theorem and Maximum a Posteriori for feature extraction, and for classification, SVM is used. After converting the images into binary values, some points are found, which was discarded using a flood fill algorithm. The plinth steps are Clipping, Segmentation, Horizontal, and Vertical Thinning Scan. Here test accuracy rate was 99.33\%. \par
In \cite{rahaman15} M. M. Rahaman et al. proposed a CNN based model.  This method normalizes the written character images and then employed CNN to classify individual characters. It does not employ any feature extraction method like previously mentioned works. The major steps are pre-processing of raw images by converting them into a gray-scale images and then training the model. In this case, test accuracy was 85.36\%. 
On another paper, we have seen the existence of auto-encoder for unsupervised pre-training through Deep CNN which consists of more than one hidden layer with 3 convolutional layers. Each layer was followed by $2 \times 2$ max-pooling layer. This scheme \cite{shopon17} was proposed by Md Shopon et al. The layers have $32\times3\times3$ number of kernels. In the same manner, the decoder has an architecture with each convolutional layer with 5 neurons, rather than 32. The ReLU \cite{relue13} activation is present in all layers. For training purposes, the model enhanced the training dataset by randomly rotating each image between 0 degrees and 50 degrees and also by shifting vertically by a random amount between 0 pixels and 6 pixels. This model was trained in 3 various setups SCM, SCMA, and ACMA. They have achieved a test accuracy of 99.50\%.
Another model \cite{akhand16}, proposed by M. A. H. Akhand et al. used pre-processing by using a simple rotation based approach to produce patterns and it also makes all images of ISI handwritten database into the same resolution, dimension, and size. CNN structure of this model has two convolutional layers with $5\times5$ sized local receptive fields and two sub-sampling layers with $2\times2$ sized local averaging areas along with input and output layers. The input layer contains 784 receptive fields for $28\times28$ pixels image. The first convolutional operation produces six feature maps. Convolution operation with kernel spatial dimension of 5 reduces 28 spatial dimension to 24 (i.e., $28+1-5$) spatial dimension. Therefore, each first level feature map size is $24\times24$. The accuracy rate of the testing is 98.45\% on ISI handwritten Bengali numerals.\par
In \cite{choudhury18}, A. Choudhury et al. proposed a histogram of oriented gradient (HOG) and color histogram for the selection of feature algorithms. Here, HOG is used as the feature set to represent each numeral item at the feature space and SVM is used to produce the output from input. The test accuracy of this algorithm is 98.05\% on CMATERDB 3.1.1 dataset (which is a benchmark Bengali handwritten numeral database created by CMATER lab of Jadavpur University, India). M. M. Hasan et al. proposed a Bengali handwritten digit recognition model based on ResNet \cite{ResNet16}. Their ensemble model from their six best models, applied on the NumtaDB dataset \cite{NumtaDB18}, achieved 99.3359\% test accuracy.   
In \cite{noor18} R. Noor et al.
proposed an ensemble model based Convolutional Neural Network for recognizing Bengali handwritten numerals. They train their model in many noisy conditions using customized NumtaDB dataset \cite{NumtaDB18}. In all cases, their model achieved more than 96\% test accuracy on this NumtaDB dataset. A recent Bengali handwritten numeral recognition work \cite{rabby19} was there, proposed by AKM  S. A. Rabby et.al. Here authors used a lightweight CNN model to classify the handwritten numeral digits. This model is trained using ISI handwritten Bengali numeral \cite{ujjwal09} and CMATERDB 3.1.1 databases with 20\% data for validation. Their work achieved validation and test accuracy as 99.74\% and 99.58\%  on ISI handwritten numerals, which is the previous best accuracy on that dataset. Their model also shows good accuracy on other benchmark datasets such as CMATERDB 3.1.1 etc. \par 
M.S. Islam et.al proposed Bayanno-Net \cite{8971167}, where they tried to recognize Bangla handwritten numerals using CNN. They worked on NumtaDB dataset \cite{NumtaDB18}. Their model achieved 97\% accuracy. A recent review article \cite{hoq2020comparative}  written by M.N. Hoq et.al given a comparative overview of existing classification algorithms to recognized Bengali handwritten numerals.
\subsection{Advancements of deep learning for image classification}
\begin{figure}[htb]
	\centering
	\includegraphics[width=0.7\linewidth]{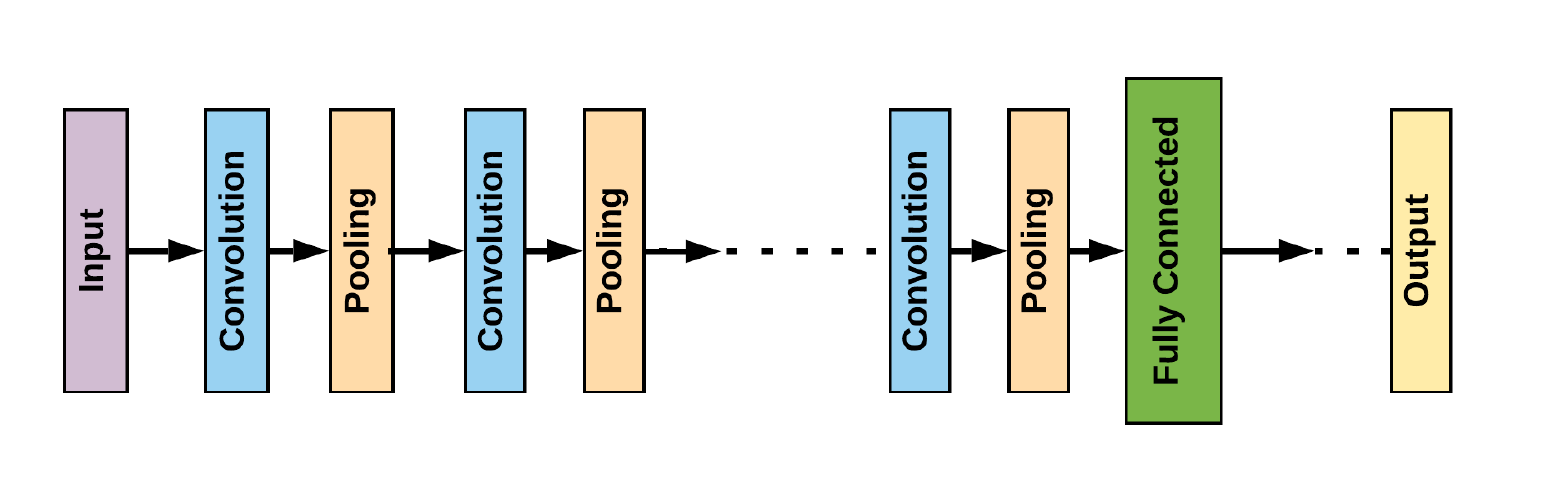}
	\caption{Building blocks of a classical CNN model \cite{farhana18}}
	\label{CNN}
\end{figure} 
After the success of AlexNet \cite{alexnet12}, a deep learning-based model for image classification,  many researchers shifted to this area of research of computer vision and pattern recognition. Therefore, many successive state-of-the-art models came within a short span of time since 2012 \cite{farhana18}, \cite{lim18}.  In this subsection, we briefly reviewed the development of deep learning especially Convolutional Neural Networks in the field of image classifications. \par    
CNN is a special type of multi-layer neural network inspired by the vision mechanism of the animal \cite{CNN19}. Hubel and Wiesel experimented and said that visual cortex cells of animal detect light in the small receptive field \cite{hubel68}. Kunihiko Fukushima got motivation from this experiment and proposed a multi-layered neural network, called NEOCOGNITRON \cite{fukushima80},  capable of recognizing visual patterns hierarchically through learning. This model is considered as the inspiration for CNN. A classical CNN model is composed of one or more blocks of convolutional and sub-sampling or pooling layer, then single or multiple fully connected layers, and an output layer function as shown in figure \ref{CNN}. The benefits of using CNN are automated features extraction, parameter sharing and many more \cite{LeNet-5}, \cite{alexnet12}, \cite{NANNI17}. Classical CNN is modified in many different ways according to target domain\cite{farhana18}, \cite{LIANG17}, \cite{BALDOMINOS18}, \cite{Farhana19}. \par
Yann LeCun et al. introduced the first complete CNN model, called LeNet-5 \cite{LeNet-5},  to classify English handwritten digit images. It has 7 layers among which 3 convolutions, 2 average pooling, 1 fully connected, and 1 output layer. They used SIGMOID function as the activation function for non-linearity before an average pooling layer. The output layer used Euclidean Radial Basis Function(RBF) for classification of MNIST \cite{yann_mnist} dataset. The weights of each layer were trained using the back-propagation algorithm \cite{rumelhart86}. AlexNet \cite{alexnet12} was the first CNN based model which won the ILSVRC challenge \cite{ILSVRC} in 2012  with a significant reduction of error. AlexNet's error rate was 16.4\% whereas the second best error rate was 26.17\%. This model was proposed by Alex Krizhevsky et al. and it is trained by using ImageNet dataset \cite{ImageNet09}, this dataset contains 15 million high resolution labeled images over 22 thousand categories. AlexNet has 11 trainable layers, and the structure is almost similar to LeNet-5, but here max-pooling used instead of the average pooling, ReLU activation in place of the SIGMOID function, softmax function in place of RBF, and $11\times11$  in-place of $5\times5$ filter size in the first layer. In addition, for the first time dropout strategy, \cite{dropout14} and GPU were used to train the model.  In \cite{ZFNet13} Zeiler and Fergus presented ZFNet which was the winner of the ILSVRC challenge in 2013. The building blocks of ZFNet is almost similar to AlexNet with few changes such as the first layer filter size is $7\times7$ instead of $11\times11$ in AlexNet. Authors of ZFNet explained how CNN works with the help of Deconvolutional Neural Networks (DeconvNet). DeconvNet is just the opposite of CNN. The error rate of ZFNet was 11.7\%. K. Simonyan and A. Zisserman proposed VGGNet \cite{VGGNet14}, which is like a deeper version of AlexNet. Here, authors used small filters $3\times3$ sizes for all layers. They have used a total 6 different CNN configurations with different weight layers. This VGGNet secured 2nd place in ILSVRC challenge in 2014 with an error rate of 7.3\% just 0.6\% more than the error rate of the winner GoogLeNet \cite{GoogLeNet15}. \par
 GoogLeNet, Going Deeper with Convolutions \cite{GoogLeNet15}, is proposed by Christian Szegedy et al. which was a research team of Google. The Structure of GoogLeNet is different from traditional CNN, it is wider and deeper than previous models but computationally efficient. Through inception architecture, multiple parallel filters with different sizes are used, and for this, problems of vanishing gradient and over-fitting were tackled. Fully connected layers are not used in GoogLeNet but the average pooling layer is used before the classifier.  This model won the ILSVRC challenge 2014 with an error rate of 6.7\%. The increasing layer could give more accuracy but will suffer from vanishing gradient problem. To tackle this problem, Kaiming He et al. from Microsoft Research proposed ResNet \cite{ResNet16}. ResNet is a very deep model where each layer has a residual block with a skip connection to the layer before the previous layer. ResNet is the winner of the ILSVRC challenge with an error rate of 3.57\% and this is a success of beyond the human level.  
Gao Hunag et al. proposed DenseNet \cite{densenet2017}, where every layer is connected to all previous layers of the model. DenseNet overcomes the vanishing gradient problem as well as it collects required features of all layers and propagates to all successive layers in feed-forward fashions for features reuse. Therefore, this model requires less number of parameters to achieve accuracy, so it is computationally efficient. Inspired by the success of ResNet, Jie Hu et al. proposed SENet \cite{SENet18} with the main focus to increase channel relationship between successive layers. SENet has added ``Squeeze-and-Excitation'' (SE) block into each block (ResNet Block), and for this, the model adaptively re-calibrates channel-wise feature responses between channels. SENet has won the ILSVRC-2017 challenge with an error rate of 2.252\%. Recently a new concept in convolutional neural networks proposed through  CapsNet \cite{NIPS2017_6975}.  Here sub-sampling concepts are abolished and classification is possible of images of different orientations.  Though the algorithm is not light for dynamic routing among capsules but it is extremely promising \cite{KWABENAPATRICK2019}. 
 \begin{figure}[htb]
	\centering
	\includegraphics[width=1.0\linewidth]{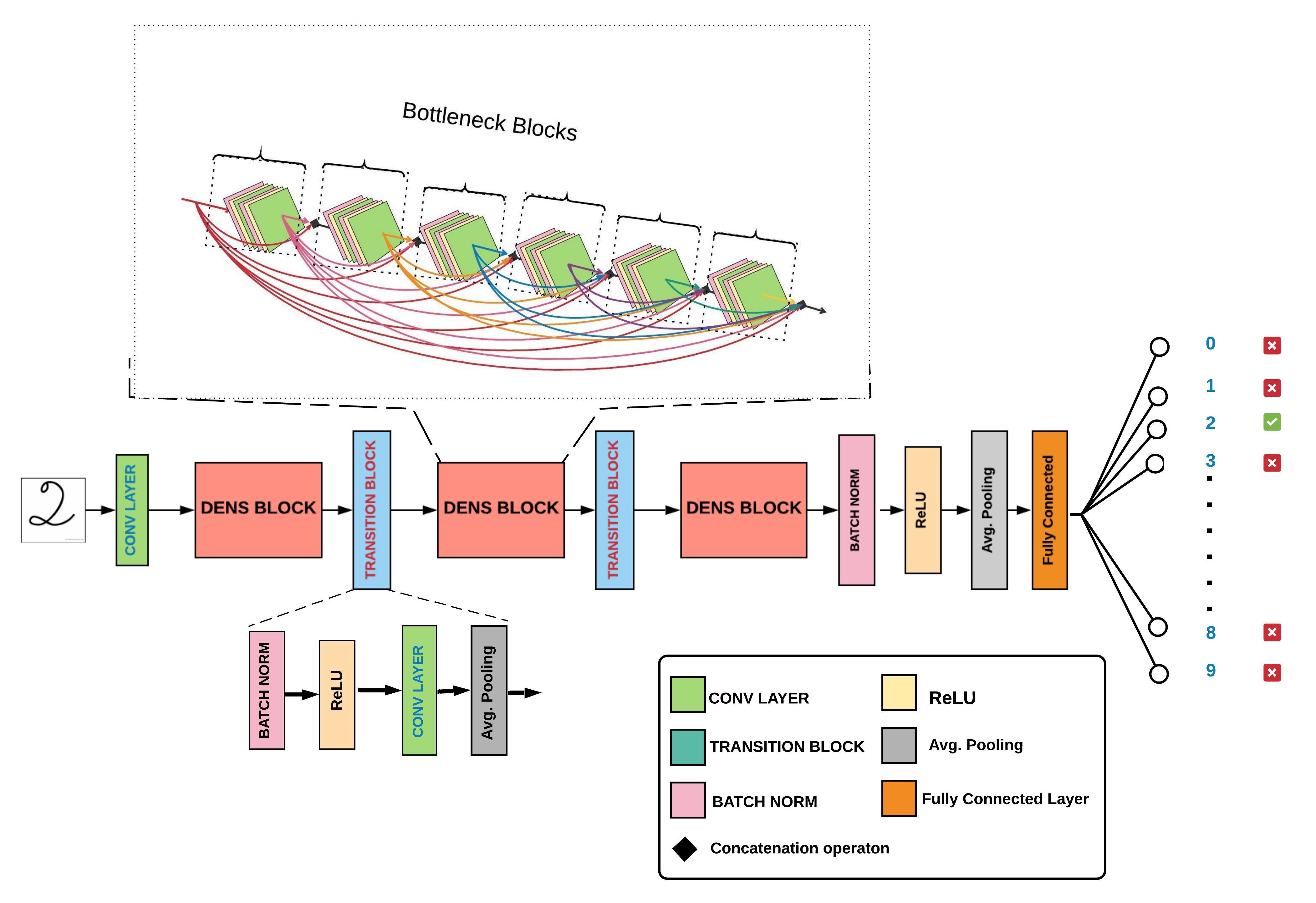}
	\caption{Structure of the BDNet }
	\label{Model}
\end{figure}
\section{BDNet Model Details} 
\label{model}
The network architecture of the BDNet model has shown in figure \ref{Model}.
The BDNet consists of three Dense Blocks and two Transition Blocks followed by Batch Normalization(BN), Rectified Linear Unit(ReLU) activation, Average Pooling(Avg. POOLING), Fully Connected(FC) Layer, Softmax Function with Output Layer.
Each Dense Block is made up of 6 bottleneck blocks. Structure of each bottleneck block is as follows: ... $\rightarrow  BatchNorm\rightarrow ReLU\rightarrow ConV2d(1\times1)\rightarrow BatchNorm\rightarrow ReLU\rightarrow ConV2d(3\times3)\rightarrow...$. The number of bottleneck blocks(NBL) per dense block has calculated using  equation \ref{Bottleneck}. 
\begin{equation}
NBL=\frac{1}{2}\floor*{\frac{n-4}{3}}
\label{Bottleneck}
\end{equation}
where $n$ is the number of layers of the network model. Dense connectivity is present among bottleneck blocks of each dense block i.e. output of each bottleneck block is forwarded to all other successive blocks for features propagation. The number of feature maps that will be forwarded depends on the growth rate, and here the growth rate is 12.
In between two dense blocks, we have used one transition block which consists of: $BatchNorm\rightarrow ReLU\rightarrow ConV.\rightarrow Avg.Polling$. To make the model compact we reduce the number of feature maps. Some hyper-parameters of the model has explained below and others which are directly related to the training in section \ref{traning}.  \\
 \textbf{Number of hidden layers and units:} It is preferably good to add more layers when the test error is no longer decreasing in existing layers. A small number of layers may lead to under-fitting, on the other hand, having more layers is usually not suitable with appropriate regularization. But adding more layers makes the model more complex and computation time will increase. After careful experiments, we have used 39 hidden layers, one fully connected(FC) layer in our model. Then we have used softmax function to output 10 classes. Model details are shown in figure \ref{Model}. Here, softmax function transforms predicted scores to predicted probability scores using the following equation \ref{Softmax}.
\begin{equation}
\hat{y_i}=\frac{e^{z_i}}{\sum_{j=1}^{10} e^{z_j}}
\label{Softmax}
\end{equation}
Where $\hat{y_i}$ denotes prediction score of  i-th digit or class. \\
 \textbf{Optimizer:} BDNet trained through back-propagation \cite{rumelhart86} using optimizer. Here, weights updated using SGD (Stochastic Gradient Descent) \cite{sgd10}. The mathematical loss function $J(w)$ as in equation \ref{Loss Function} and it is basically the difference between the updated internal parameters of a model which are used for computing the values ($y_i$) from the set of  inputs ($x_i$) used in the model and the desired output ($\hat{y_i}$).  
\begin{eqnarray}
J(w)&=&\frac{1}{n}\sum_{i=1}^{n}J_i(w) \nonumber \\ 
&=&\frac{1}{n}\sum_{i=1}^{n}(y_i-\hat{y_i})^2
\label{Loss Function}
\end{eqnarray}
The working flow of optimizer for BDNet is as follows:\\
Step 1: Initialization of the vector of parameters $w$ and learning rate $\eta$.\\ 
Step 2: Repeat until an approximate minimum is found:\\
\indent Step 2.1: Randomly shuffle items in the training set.\\
\indent Step 2.2: for $i = 1$ to $n$ do:\\
\indent \indent \indent \indent $w = w-\eta\nabla J_i(w)$\\
BDNet used random order of the training dataset. As the coefficients are updated after each training data-sample, so the updates, as well as the lost function will be randomly jumping all over the place. By this randomized updates to the coefficients, it reduces random walk and avoids distraction.\\
\textbf{Activation Function:} We have used Rectified Linear Unit (ReLU) \cite{relue13} as activation function. ReLU function as in equation \ref{ReLu} works for non-linearity.
\begin{equation}
f(x)=\max(0, x)
\label{ReLu}
\end{equation}
Here $x$ denotes the value of a pixel. ReLU removes negative values from an activation map by setting them to zero. It increases the nonlinear properties of the decision function and removes the chances of vanishing gradient of BDNet without affecting the receptive fields of the convolution layer. 
\section{Dataset and Preprocessing of the Dataset} 
\label{dataset}
\begin{figure}[htb]
	\centering
	\includegraphics[width=.90\linewidth]{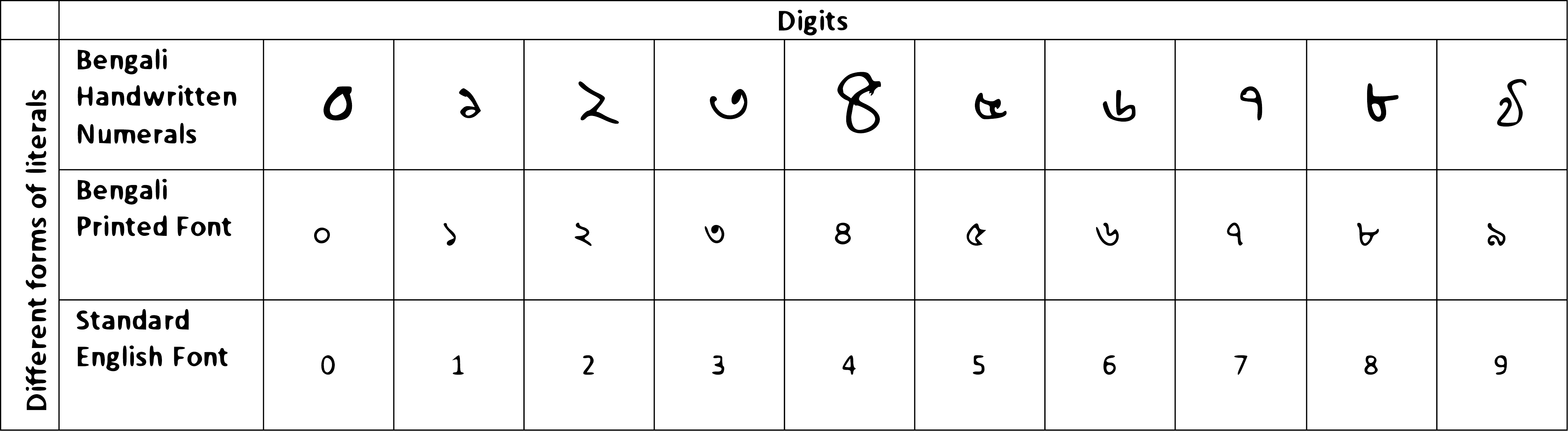}
	\caption{Typical Bengali handwritten numeral digits and corresponding printed values.}
	\label{sample}
\end{figure} 
The Bengali language is mainly derived from the Brahmi script and Devanagari script in the 11th Century AD. The structural view of each character and numeral of this language are very complex.  So, train a model using the Bengali digit is more difficult compared to the English numeral digit as the English digits have a less complex structure. In addition, English numerals datasets are easily available in terms of quantity and quality such as MNIST \cite{yann_mnist} but it is not easy for Bengali numeral datasets. Bengali digit also has some high similarity features for different numerals such as numeral 1 (in Bengali) and numeral 9 (in Bengali) has high similarity features, similarly numeral 5 (in Bengali) and 6 (in Bengali) has high similarity features. The typical Bengali handwritten numerals and corresponding printed values as shown in figure \ref{sample}.
 \begin{figure}[htb]
	\centering
	\includegraphics[width=.65\linewidth]{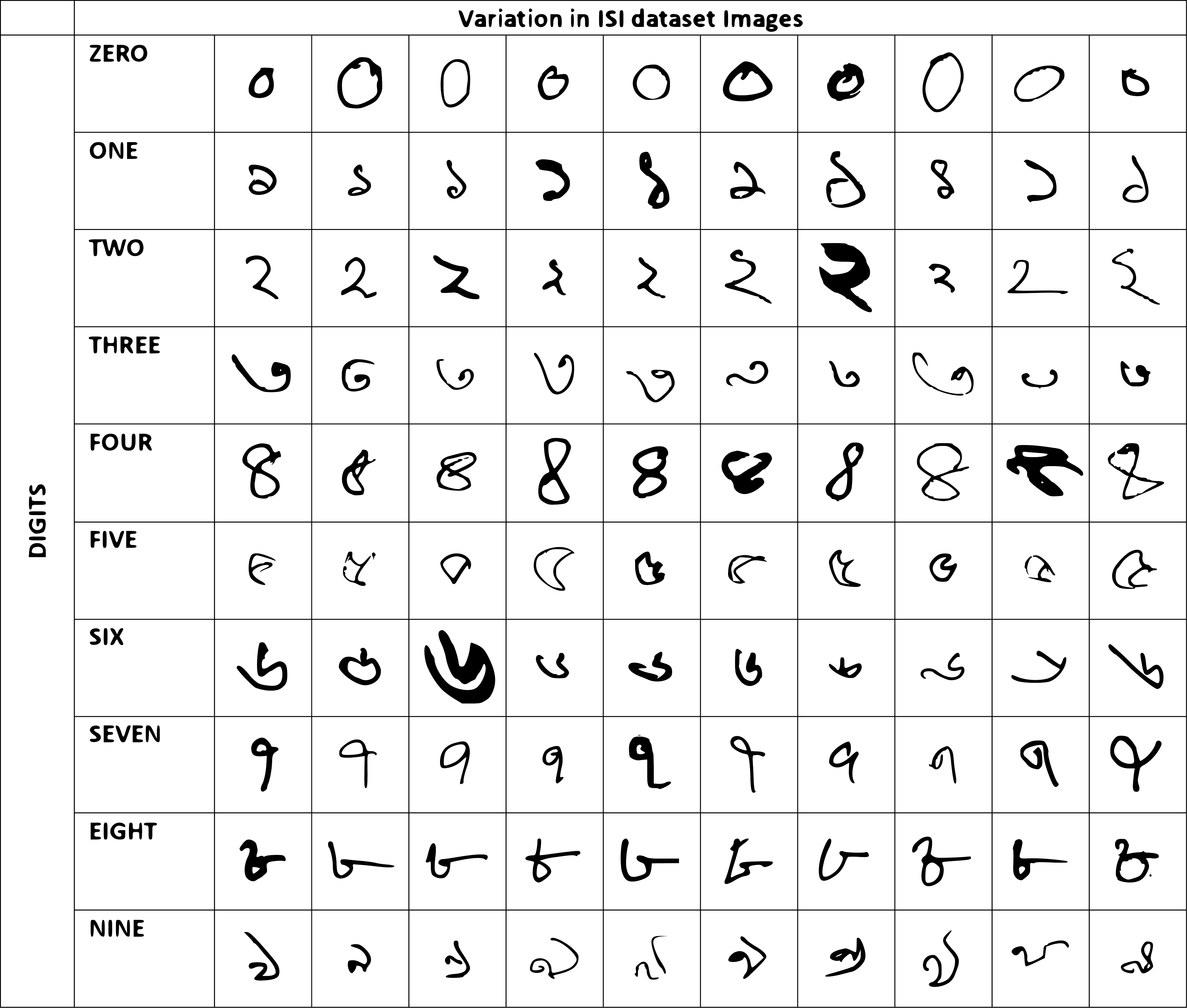}
	\caption{Sample handwritten ISI numeral image data}
	\label{Sample ISI numeral image data}
\end{figure}
\subsection{Used Dataset}The ISI Bengali numeral off-line handwritten dataset \cite{ujjwal09} is one of the largest popular datasets of handwritten Bengali numerals. This dataset consists of 23392 black and white image data written by 1106 persons collected from postal mail and job application forms. Among these 23392 data, 19392 are training data and 4000 are testing data. The entire dataset represents 10 classes for numeral 0 to 9. Some typical data items of this dataset shown in figure \ref{Sample ISI numeral image data}.

\subsection{Preprocessing of the Datasets }
 \begin{figure}[htb]
 	\minipage{0.47\textwidth}
 	\centering
	\includegraphics[width=.9\linewidth]{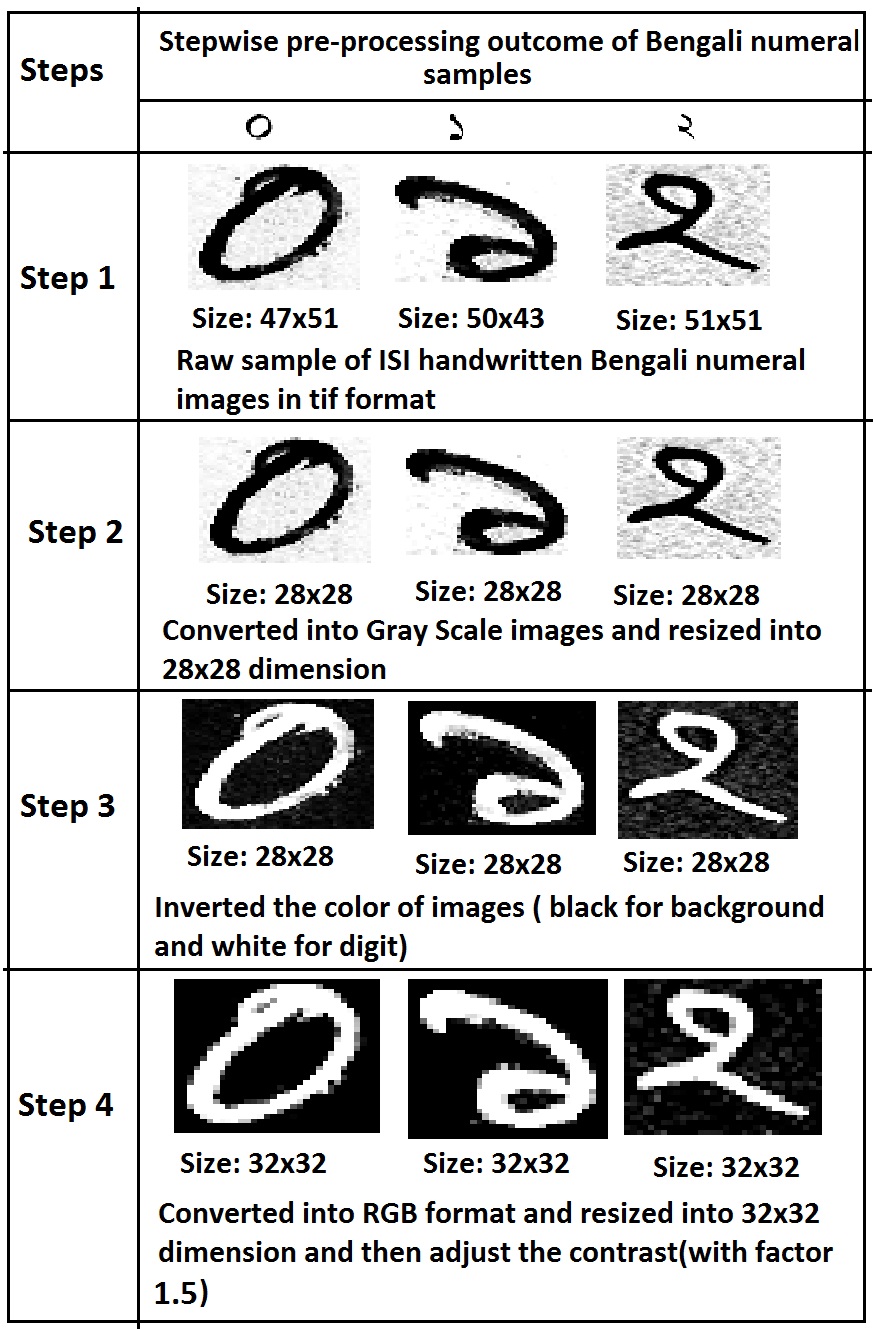}
	\caption{Pre-processing steps}
	\label{pre-processing steps}
	\endminipage\hfill
	\minipage{0.495\textwidth}
	\centering
		\includegraphics[width=.9\linewidth]{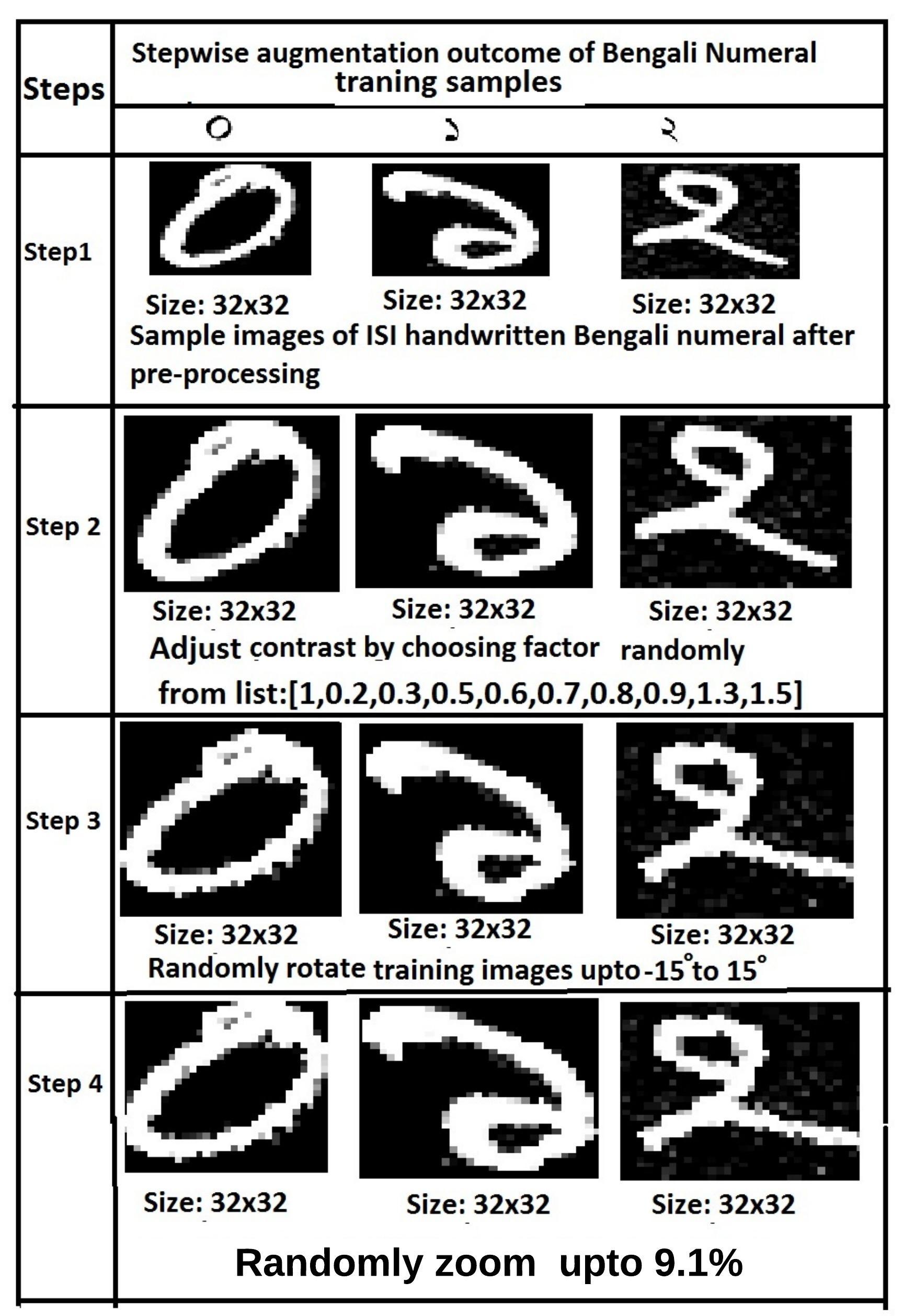}
		\caption{Data augmentation steps.}
		\label{augmentation steps}
	\endminipage\hfill
\end{figure}
As mentioned we have used the ISI Handwritten numeral dataset to train the BDNet. But the data items that we have chosen for this task is very untidy and cannot be used directly for our purpose. All the data were raw images in \textbf{.tif} format of different sizes. First, we have converted the raw images into gray-scale images of size $28\times28$, then inverted the colors in a way that the background became black and the font became white. After that gray-scale images are converted to RGB images of size $32 \times 32$ for better feature extraction using 3 channels. For the convenience to use the BDNet, we have created a CSV file to access the data samples. Figure \ref{pre-processing steps} is showing  steps of preprocessing, and how converted data looks different from actual data after the preprocessing. Distribution of entire ISI handwritten numerals database as shown in table \ref{table of data distribution}. Among the training datasets, 10\%  of data is used for 10-fold cross-validation. As we have used unorthodox data augmentation techniques, so the size of the dataset virtually becomes large, and for large dataset,  10-fold cross-validation worked fine. 
\begin{table}[htb]
	\caption{Used Dataset Distribution}
	\label{table of data distribution}
	\begin{center}
		\begin{tabular}{|c|c|c|}
			\hline\noalign{\smallskip}
			\textbf{Digit} & \textbf{Training Sets} & \textbf{Test Set}\\
			\hline \noalign{\smallskip} 
			0 & 1933 & 400\\
			1 & 1945 & 400\\
			2 & 1945 & 400\\
			3 & 1956 & 400\\
			4 & 1945 & 400\\
			5 & 1933 & 400\\
			6 & 1930 & 400\\
			7 & 1928 & 400\\
			8 & 1932 & 400\\
			9 & 1945 & 400\\
			\hline
		\end{tabular}
	\end{center}
\end{table} 
\subsection{Own dataset}
\label{own dataset}
We have also created our own dataset of 1000 images. Among these 1000 images, 100 images per digit are there for each Bengali numeral digit from digits zero to nine. This dataset is created by 5 laboratory members of this work with the help of some students. It has done by writing the digits in standard pages using black or blue pens. Then we have scanned the written digits using the mobile phone camera. Datasets are created with the focus to make it as natural as the common people write the Bengali numerals in their daily life. Each image of the dataset is then set to $28\times28$ pixels. We have used this dataset only for testing to check the generic performance of the trained BDNet.  
\section{Training Details}
\label{traning}
We have started our experiment by training the model using the preprocessed labeled dataset mentioned above. Before describing the training details, we have presented the system environment and resources used for our work in table \ref{System Specification}.
\begin{table}[htb]
	\caption{Used system specifications}
	\label{System Specification}
\begin{center}
		\begin{tabular}{|c|l|}
			\hline
			\textbf{Resources} & \textbf{Specifications}\\ 
			\hline \noalign{\smallskip}
			CPU & $Intel^R$ Xeon$^R$ CPU $ @ 2.3GHz$ with 45 MB Cache \\ \hline 
			RAM & 12.72 GB available \\ \hline 
			DISK & 1 TB (Partially used) \\ \hline 
			GPU & $1\times $ Nvidia Tesla T4 having 2560 CUDA cores,\\ & 16GB(14.72GB available) GDDR6 VRAM \\ \hline 
			Languages \& Packages & Python with PyTorch\cite{pytorch17}\\ \hline 
			Training  \& Validation Time & 37.68 Seconds per Epoch\\
			\hline
		\end{tabular}
\end{center}
\end{table}
The design of BDNet is inspired by the state-of-the-art algorithm DenseNet. A new set of hyper-parameters here established and set the required value for each required hyper-parameter is a very difficult task. It could be done by trial and error method with careful observation of the pattern of the data as well as by some mathematical analysis. In a similar fashion, we have done hyper-parameter tuning of required hyper-parameters of BDNet. Some are mentioned in section \ref{Model} others are described below. \\
 \textbf{Number of Epochs:} As it is the number of time the entire training dataset passes through the model network. So, we could increase the number of epochs until the training error becomes small and the validation error is noticeable. For BDNet model, the number epoch was set to 150 but the model converges around 125 epochs, and it took 37.68 seconds per epoch to train and validate simultaneously(in our system mentioned in table \ref{System Specification}).\\
 \textbf{Weight initialization:} We have initialized the weights with small random numbers (between 0 and 1) to prevent dead neurons, but not too small numbers to avoid zero gradients. Uniform distribution usually works very well. Here, we  have used $seed(1)$ as a python function to initialize the weights randomly. \\
\textbf{Batch size:} Mini-batch is usually preferable in the place where a dataset is very large. It is usually used to create a partition in between the dataset. Batch size doesn't contribute much to the precision but helps to control the training speed. We have used the batch size as 32 and the batch size was 64 for testing.\\
\textbf{Learning rate:} In BDNet we have used the learning rate as 0.009 and after 80 epoch it has been changed as in equation \ref{lr}.
\begin{equation}
\eta=(\mbox{Intial}\ \eta) \times (0.15)
\label{lr}
\end{equation}
\textbf{Weight decay:} This is another hyper-parameter tuning where each step's current weights($W$) are multiplied by a number slightly less than 1. Weight decay is a regularization term that prevents growing the number of parameters into a large number. It is updated as  equation \ref{Weight Decay}
\begin{equation}
W_i = W_i-\eta\frac{\partial J }{\partial W_i} - \eta\lambda W_i
\label{Weight Decay}
\end{equation}
Where $J$ is the current loss, $\eta$ is the learning rate and $\lambda$ is the weight decay. After an experiment with several values, finally, the value of the weight decay was set \textbf{\emph{1e-5}}. \\
\textbf{Momentum:} The concept of momentum is that previous changes in the weights should influence the current direction of movement in weight space.  Sometimes these weights changes stuck in a local minimum. To avoid these local minima, BDNet has used  momentum in the objective function, which is a value between 0 and 1 that increases the size of the steps taken towards the minimum by trying to jump from a local minimum. Here the momentum value set is 0.9, and for this reason, speed and accuracy improve.\\
\textbf{Dropout for Regularization:} Few techniques ideas make deep learning popular and usable, and the dropout for regularization \cite{dropout14} is one of them.  Dropout is used for BDNet to avoid over-fitting during training. The method simply drops out some neurons randomly in the neural networks in each iteration of training according to a threshold probability. Here we used the dropout threshold probability as 0.09 which is a small probability i.e. only 9\% neuron dropout in each epoch.\\

\textbf{Data Augmentation:} 
Data augmentation is one of the important parts which gives more versatility to extracted features and helps to train the deep learning model more accurately.  Here, we have used data augmentation as the size of the dataset is not as large as required. But the idea has been used with a slightly non-traditional way as shown in figure \ref{augmentation steps}.  As an image of the handwritten numeral digits have some problems as we can not crop or rotate largely. Here, we did augmentation on the training set for each epoch as follows: adjusted the contrast of training samples by choosing the adjusting factor randomly from the list $[1,0.2,0.3,0.5,0.6,0.7,0.8,0.9,1.3,1.5]$, random rotation of training images from -15 to +15 degrees, and random zooming up to 9.1\%. For this type of data augmentation, a slightly new dataset is passed through the network in every iteration or epoch. \\
\textbf{Cross-Validation:} To train BDNet 10-fold random cross-validation is used. We used 10\% of training data only for 10-fold cross-validation to validate the model during training for generalization without over-fitting. After one epoch, the training set is re-sampled with 10-fold cross-validation. The cross-validation result is mentioned in section \ref{result}.  
\section{Result Analysis}
\label{result}
 \begin{figure}[htb]
	\centering
	\includegraphics[width=.99\linewidth]{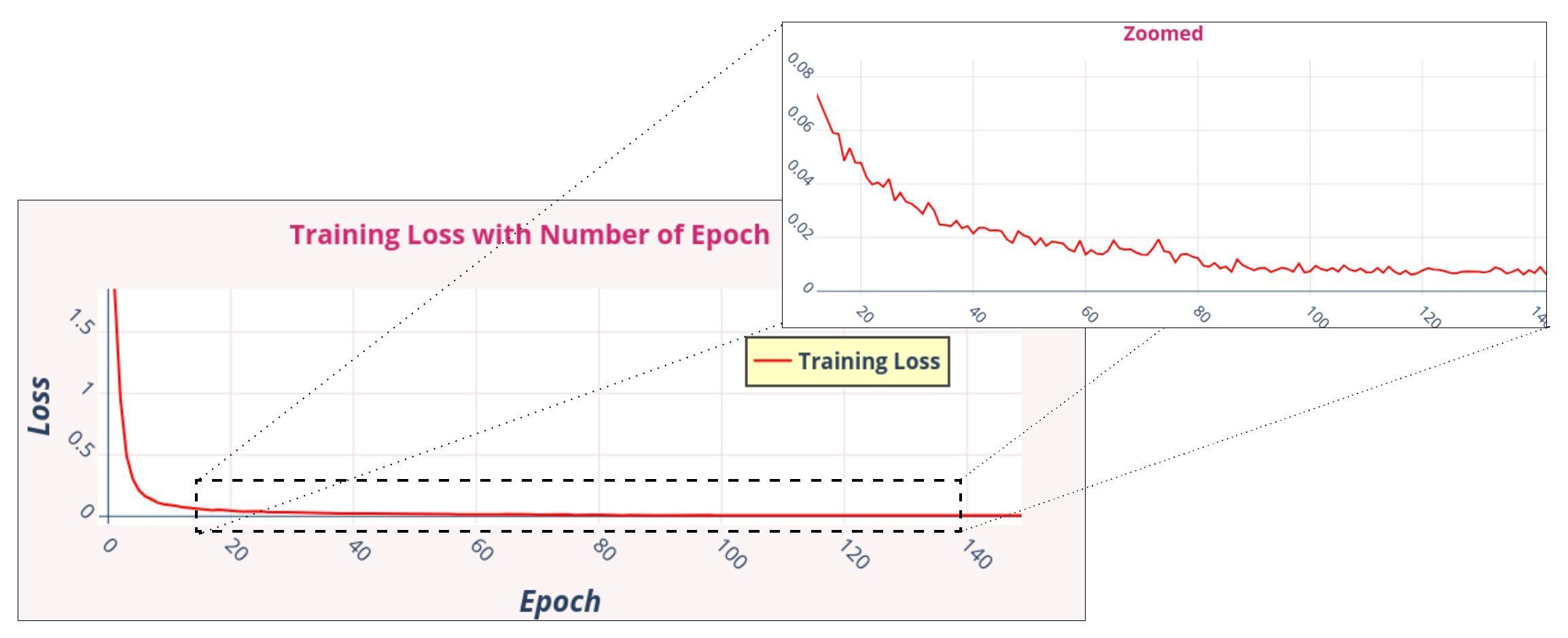}
	\caption{Number of epoch vs training loss}
	\label{Epoch vs training}
\end{figure} 
 \begin{figure}[htb]
	\centering
	\includegraphics[width=.95\linewidth]{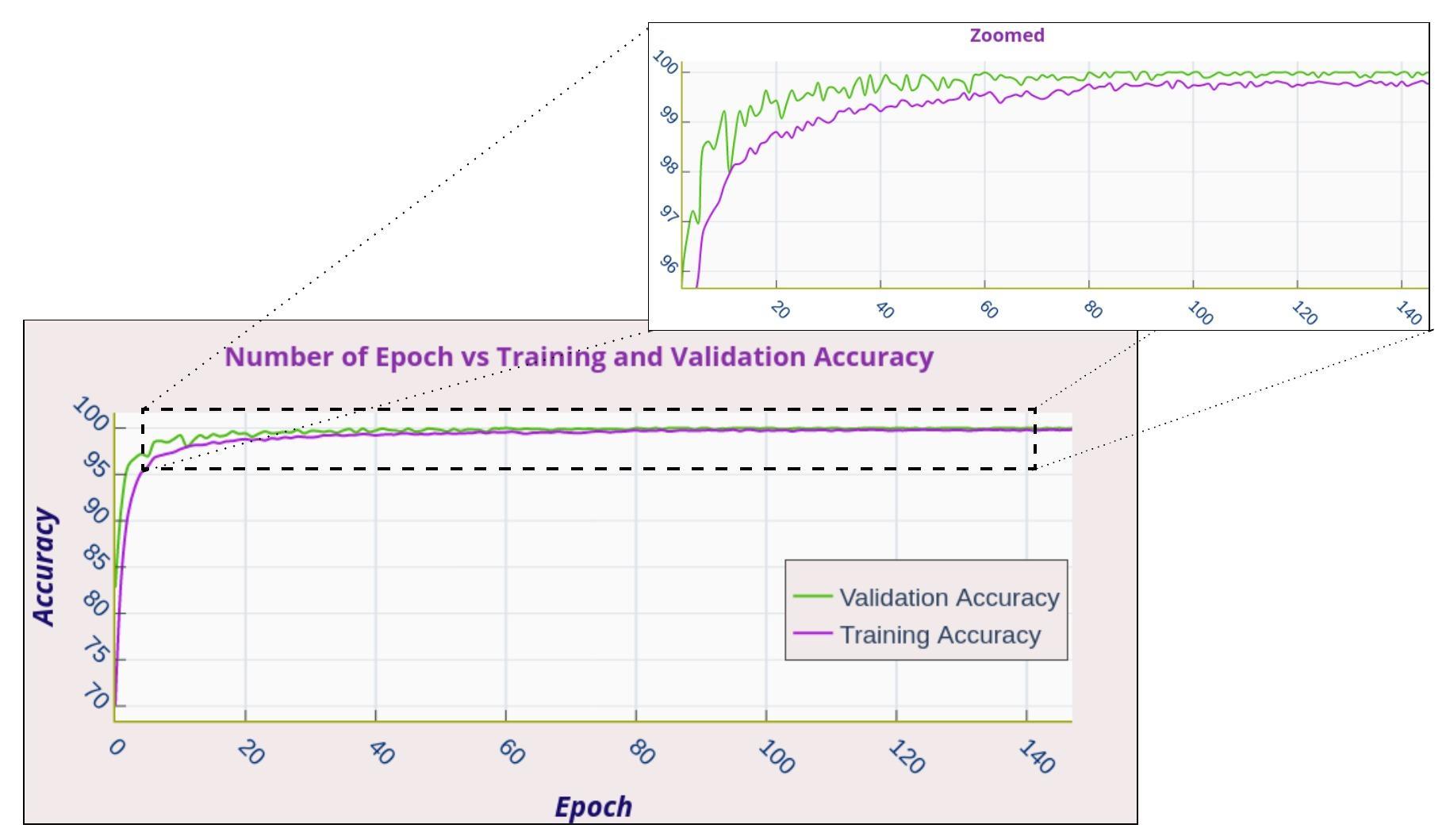}
	\caption{Number of epoch vs Training and Validation Accuracy}
	\label{train_valid}
\end{figure}
It has mentioned that the BDNet is trained using a pre-processed ISI handwritten Bengali numeral database with data augmentation and 10-fold cross-validation. BDNet is tested using the test dataset from the same database mentioned in section \ref{dataset}. The trained model also tested using our own dataset described in that section. The following subsections are showing some results of the BDNet, found during training, validation and testing.
\subsection{Number of Epoch vs Training Loss} 
At first, when the training started, the amount of error or training loss was very high and the value of error rate was between 0.95 to 1.88. But with the increasing number of training epochs, the value of training loss decreased drastically and later it slowed down as shown in figure \ref{Epoch vs training}.  After 120 epochs, the error rate became very small almost between 0.008 to 0.006.
 \begin{figure}[htb]
	\centering
	\includegraphics[width=.98\linewidth]{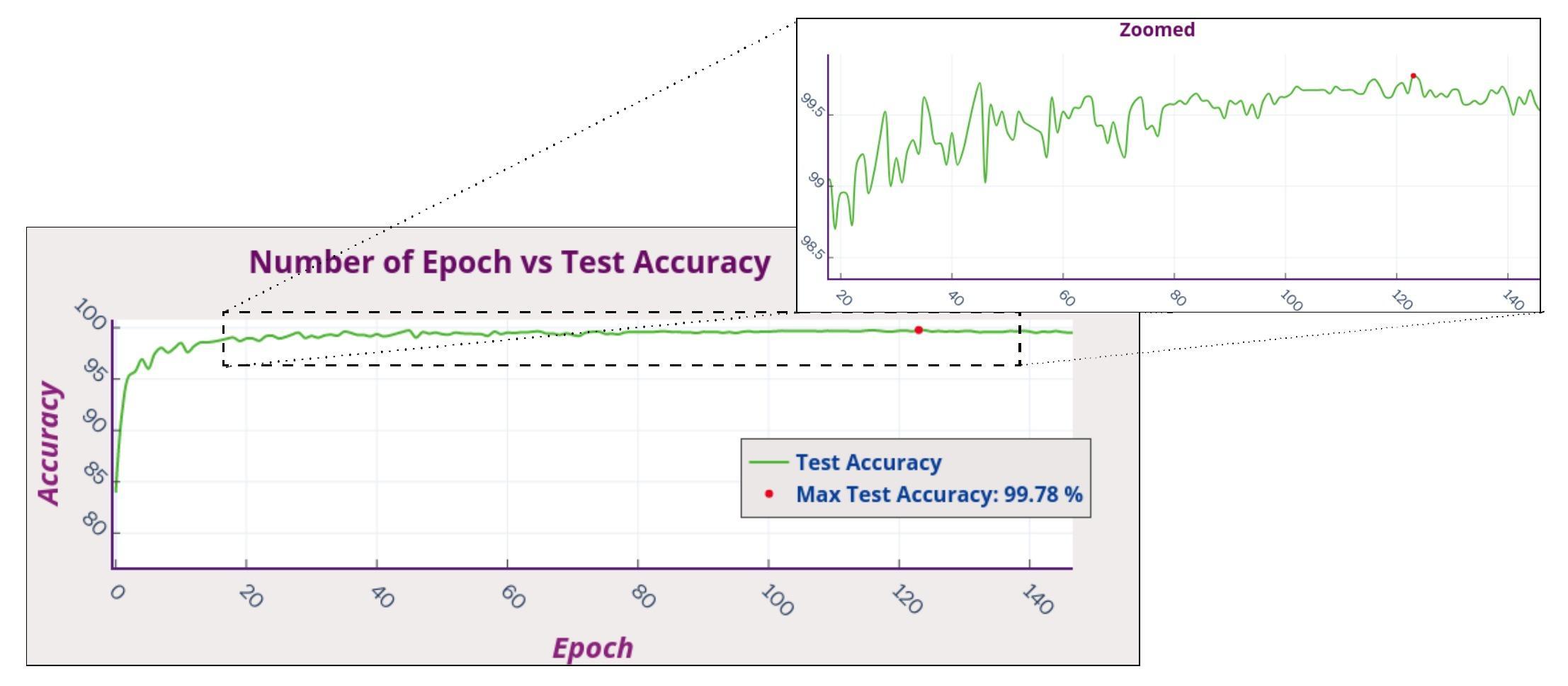}
	\caption{Number of epoch vs Test Accuracy}
	\label{testing}
\end{figure}
\subsection{Training and Validation Accuracy}
The training and validation accuracy has been observed simultaneously. Here validation are used to reduced over-fitting the model during training. As we can see in figure \ref{train_valid}, the increasing rate of accuracy was very high during the initial training period and it gradually became very low. After 125 epochs, it was almost saturated. We can also see from this figure that the training accuracy was mostly dominated by validation accuracy, and it is happened because of data augmentation. Maximum training accuracy was recorded \textbf{99.82\%} after epoch number 118, and maximum validation accuracy was recorded \textbf{100\%} in many times such as after 97th, 107th, 118th, 120th epoch. Although validation accuracy(test accuracy is more vital for any task-oriented model) is not sufficient for any model but here great validation recorded. This was achieved because of data augmentation and preprocessing.    
\subsection{Test Accuracy:} It is mentioned that BDNet has achieved record-breaking test accuracy on ISI Bengali handwritten numeral test dataset. BDNet achieved  99\% test accuracy after 45 epochs, and at 125th epoch, it achieved \textbf{99.78}\% test accuracy as shown in figure \ref{testing}, which is the current best result on the ISI handwritten numeral dataset. Deatils of the testing results described in subsection \ref{CM}.

\subsection{Analysis through Confusion Matrix}
\label{CM}
Testing result of the BDNet on the test dataset of ISI handwritten Bengali numeral can be presented in the confusion matrix as shown in table \ref{Confusion Matrix1}. 
\begin{table}[htb]
	\caption{Confusion matrix of the test result of the ISI handwritten Bengali numerals test dataset.}
	\label{Confusion Matrix1}
	\begin{center}
		\begin{tabular}{|c|c|r|r|r|r|r|r|r|r|r|r|r|}\cline{1-13}
			& & \multicolumn{11}{|c|}{Predicted Class}\\
			\cline{1-13}
   	       & \rotatebox[]{45}{Numerals} & 0 & 1 & 2 & 3 & 4 & 5 & 6 & 7 & 8 & 9 & \rotatebox[]{305}{Accuracy(\%)} \\ 
   	       \cline{2-13}\hline
           \multirow{10}{*}{\rotatebox[]{90}{Actual Class}}
           & 0 & 398 & 0 & 0 & 1 & 0 & 0 & 0 & 1 & 0 & 0 & 99.50 \\ \cline{2-13}
           & 1 & 0 & 397 & 1 & 0 & 0 & 0 & 0 & 0 & 0 & 2 & 99.25\\ \cline{2-13}
           & 2 & 0 & 0 & 399 & 0 & 0 & 1 & 0 & 0 & 0 & 0 & 99.75 \\ \cline{2-13}
           & 3 & 0 & 0 & 0 & 400 & 0 & 0 & 0 & 0 & 0 & 0 & 100.00\\ \cline{2-13}
           & 4 & 0 & 0 & 0 & 0 & 400 & 0 & 0 & 0 & 0 & 0 & 100.00\\ \cline{2-13}
           & 5 & 0 & 0 &  0 & 0 & 1 & 398 & 1 & 0 & 0 & 0 & 99.50\\ \cline{2-13}
           & 6 & 0 & 0 &  0 & 0 & 0 & 0 & 400 & 0 & 0 & 0 & 100.00 \\ \cline{2-13}
           & 7 & 0 & 0 & 0 & 0 & 0 & 0 & 0 & 400 & 0 & 0 & 100.00 \\ \cline{2-13}
           & 8 & 0 & 0 & 0 & 0 & 0 & 0 & 0 & 0 & 400 & 0 & 100.00 \\ \cline{2-13}
           & 9 & 0 & 0 & 0 & 0 & 0 & 1 & 0 & 0 & 0 & 399 & 99.75 \\ 
           \hline 
		\end{tabular}
	\end{center}
\end{table}
Clearly, we can see that among 10 numeral digits of 4000 test images, only 9 were wrongly predicted or classified. Among 400 test images of the digit 0, one is incorrectly predicted as 3, and one is 7. Similarly, 3 images of the digit 1, 1 image of the digit 2, 2 images of the digit 5, and 1 image of the digit 9 were predicted incorrectly. The details of wrong predictions are shown in the confusion matrix in table \ref{Confusion Matrix1}. The \textbf{F1} score for this confusion matrix is 0.99775. 

A total of 9 images are wrongly classified among all the test images of the dataset which are shown in table \ref{wrongly callssified images}. In the confusion matrix, it has shown that which wrongly predicted image is classified into which class.  After careful observation of the patterns of these 9 images, the possible reasons behind these wrong classification could be understood. 
\begin{table}[htb]
	\caption{Wrongly Classified Images}
	\label{wrongly callssified images}
	\begin{center}
		{\renewcommand{\arraystretch}{1.4}
		\begin{tabular}{|c|c|c|}
			\hline
			\textbf{Digit Images} & \textbf{Actual Class} & \textbf{Predicted Class}\\ \noalign{\smallskip}
			\hline
			
			\includegraphics[height=.35 cm]{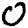}  & 0 & 3\\ \hline
	     	\includegraphics[height=.35 cm]{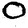} & 0 & 7\\ \hline
	     	\includegraphics[height=.30 cm]{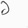} & 1 & 2\\ \hline
	     	\includegraphics[height=.35 cm]{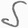} & 1 & 9\\ \hline
	        \includegraphics[height=.35 cm]{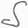} & 1 & 9\\ \hline
	        \includegraphics[height=.35 cm]{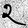} & 2 & 5\\ \hline
	        \includegraphics[height=.35 cm]{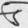} & 5 & 4\\ \hline
	        \includegraphics[height=.35 cm]{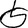} & 5 & 6\\ \hline
	        \includegraphics[height=.35 cm]{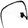} & 9 & 5\\ \hline
		\end{tabular}
	}
	\end{center}
\end{table}
\subsection{Standard Deviation of test results}
To find how spread out the test accuracy in case of training the model multiple times with the same hyper-parameter configurations, we calculate the standard deviation ($\sigma$) by using equation \ref{sd}.
\begin{equation}
\sigma=\sqrt{\frac{1}{N}{\sum_{i=1}^{N}{ {(x_i - \mu)}^2}}}  
\label{sd}
\end{equation}
Where $\mu$ is the mean of $N$ values.\\
Now, we run our model 5 times with the same hyper-parameter configurations and the achieved test accuracy has shown in table \ref{test accuray in every cases} and we get $\mu=99.75$. So finally we get the standard deviation ($\sigma$) of test accuracy as \textbf{$0.015$}. Codes along with training details of all those 5 cases are available at:\\ \href{https://github.com/Sufianlab/BDNet}{https://github.com/Sufianlab/BDNet}.

\begin{table}[htb]
	\caption{Test accuracy of our model in 5-classes cases when we run the model with same hyper-parameter configurations.}
	\label{test accuray in every cases}
	\begin{center}
		\begin{tabular}{|c|c|r|}\cline{1-3}
			& & \multicolumn{1}{|c|}{Test Accuracy (\%)}\\ 
			\cline{2-3}\hline
			\multirow{5}{*}{\rotatebox[]{90}{Cases}}
			
			& 1 & 99.75 \\ \cline{2-3}
			& 2 & 99.73  \\ \cline{2-3}
			& 3 & 99.75 \\ \cline{2-3}
			& 4 & 99.78 \\ \cline{2-3}
			& 5 & 99.75 \\ \cline{2-3}			
			\hline 
		\end{tabular}
	\end{center}
\end{table}
\subsection{Comparison of Test Results with Base-line-models}
\begin{figure}[htb]
	\centering
	\includegraphics[width=.77\linewidth]{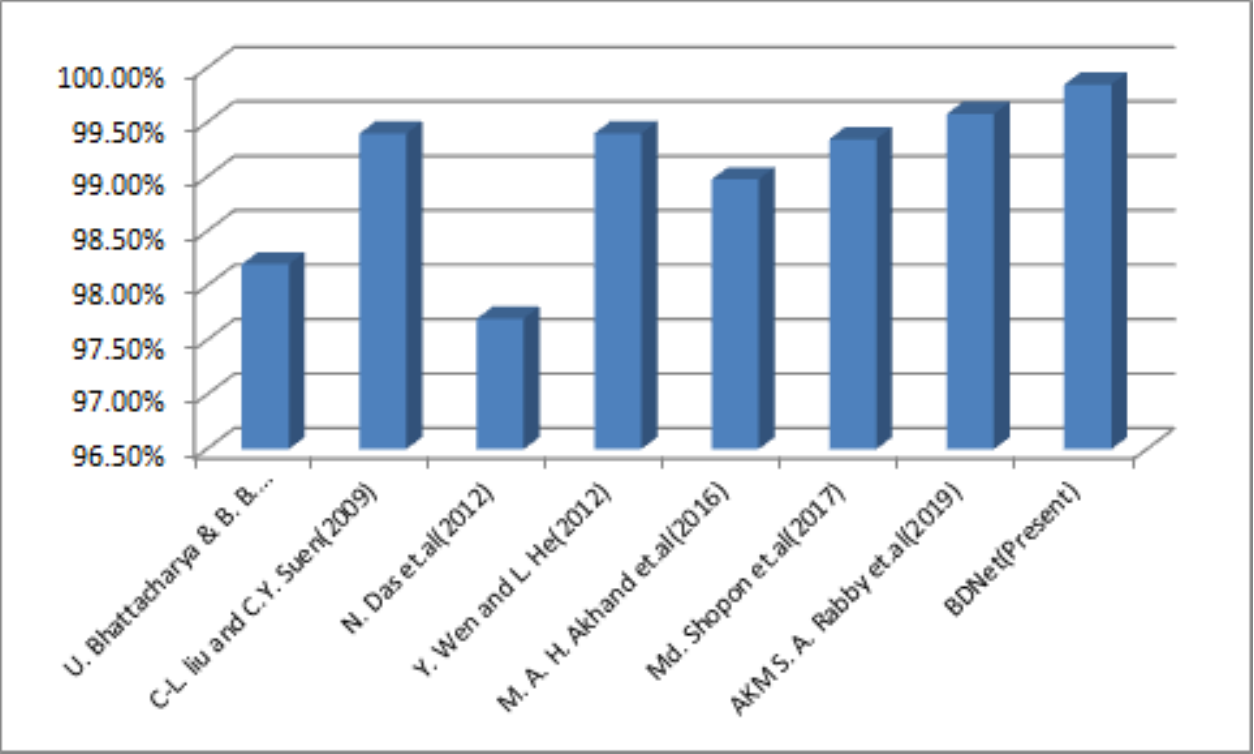}
	\caption{Comparison of notable models and corresponding test accuracy.}
	\label{comparison fig}
\end{figure}

The most notable models proposed by researchers for Bengali handwritten numerals recognition has been discussed in subsection \ref{exixting works}. We have compared BDNet with some notable models which are also worked on this benchmark dataset \cite{ujjwal09}. Before BDNet, the previous two best models \cite{liu09} and \cite{rabby19} achieved the test accuracy of 99.40\% and 99.58\%(authors of \cite{rabby19} claimed it) respectively whereas BDNet achieved 99.78\%. All the notable models and corresponding test accuracies are shown in  table \ref{comparisons table}. Graphical comparison of said models has shown in
figure \ref{comparison fig} where X-axis presents the models and Y-axis showed the corresponding test accuracy in the benchmark dataset \cite{ujjwal09}.
\begin{table}[htb]
	\caption{Notable Bengali handwritten numerals recognition models and corresponding test accuracy in the benchmark dataset \cite{ujjwal09}.}
	\label{comparisons table}
	\begin{center}
		\begin{tabular}{|l|c|}
			\hline
			\textbf{Models} & \textbf{Test accuracy}\\
			\hline \noalign{\smallskip}
			U. Bhattacharya \& B. B. Choudhury(2009)\cite{ujjwal09}	& 98.20\% \\ \hline
			C-L. Liu \& C.Y. Suen(2009) \cite{liu09} &	99.40\% \\ \hline
			N. Das et.al(2012) \cite{nibaran12} & 97.70\% \\ \hline
			Y. Wen and L. He(2012) \cite{wen12} & 99.40\% \\ \hline
			M. A. H. Akhand et.al(2016) \cite{akhand16} &	98.98\% \\ \hline
			Md. Shopon et.al(2017) \cite{shopon17} & 99.35\% \\ \hline
			AKM S. A. Rabby et.al(2019) \cite{rabby19}  & 99.58\% \\ \hline
			BDNet(Proposed in this paper) & \textbf{99.78}\%\\ \hline
		\end{tabular}
	\end{center}
\end{table}
\subsection{Test Result on Our Own Dataset}
Our own test dataset is described in subsection \ref{own dataset} which has 10 classes with 100 images per class. This test dataset is not used during training for evaluating the generalization capability of the trained BDNet on a new dataset. As a result, the BDNet has got 98.80\% test accuracy. The entire result has been shown in the confusion matrix mentioned in table \ref{Confusion Matrix2}.
 \begin{table}[htb]
 	\caption{Confusion matrix of the test result on own test dataset}
 	\label{Confusion Matrix2}
 	\begin{center}
 		\begin{tabular}{|c|c|r|r|r|r|r|r|r|r|r|r|r|}\cline{1-13}
 			& & \multicolumn{11}{|c|}{Predicted Class}\\
 			\cline{1-13}
 			& \rotatebox[]{45}{Numerals} & 0 & 1 & 2 & 3 & 4 & 5 & 6 & 7 & 8 & 9 & \rotatebox[]{305}{Accuracy(\%)} \\ 
 			\cline{2-13}\hline
 			\multirow{10}{*}{\rotatebox[]{90}{Actual Class}}
 			& 0 & 100 & 0 & 0 & 0 & 0 & 0 & 0 & 0 & 0 & 0 & 100  \\ \cline{2-13}
 			& 1 & 0 & 100 & 0 & 0 & 0 & 0 & 0 & 0 & 0 & 0 & 100\\ \cline{2-13}
 			& 2 & 0 & 0 & 100 & 0 & 0 & 0 & 0 & 0 & 0 & 0 & 100\\ \cline{2-13}
 			& 3 & 0 & 0 & 0 & 98 & 0 & 0 & 1 & 0 & 1 & 0 & 98\\ \cline{2-13}
 			& 4 & 0 & 0 & 0 & 0 & 100 & 0 & 0 & 0 & 0 & 0 & 100 \\ \cline{2-13}
 			& 5 & 1 & 0 &  0 & 0 & 0 & 95 & 3 & 1 & 0 & 0 & 95\\ \cline{2-13}
 			& 6 & 0 & 0 &  0 & 0 & 0 & 0 & 100 & 0 & 0 & 0 &  100\\ \cline{2-13}
 			& 7 & 0 & 0 & 1 & 1 & 0 & 0 & 0 & 98 & 0 & 0 &  98\\ \cline{2-13}
 			& 8 & 0 & 0 & 1 & 0 & 0 & 0 & 1 & 0 & 98 & 0 & 98 \\ \cline{2-13}
 			& 9 & 0 & 1 & 0 & 0 & 0 & 0 & 0 & 0 & 0 & 99 & 99\\ 
 			\hline 
 		\end{tabular}
 	\end{center}
 \end{table}    
\section{Conclusion} 
\label{conclusion}
Aim of the work was to propose a practical task-oriented model to recognize Bengali handwritten numerals with good accuracy, and through this paper, we propose BDNet.  BDNet is a densely connected deep CNN model for handwritten Bengali numeral recognition through image classification. The BDNet is trained using ISI Bengali handwritten numerals dataset and the trained model has achieved a new benchmark accuracy on a test dataset. The trained BDNet also tested using our own dataset to see the generalization of the model, where a good result has been found. 
\section*{Acknowledgment}
First of all, the authors of the BDNet are thankful to CVPR Unit, Indian Statistical Institute, Kolkata, for providing the dataset for academic research. The authors also thankful to the editor and reviewers for mentioned some comments on initial submission to implement in the revised version. 

\small{\bibliography{sufianDLBib}}

\end{document}